\documentclass[10pt,a4paper,twoside]{article}

\usepackage{PanAfriConAI22}

\renewcommand{\doctitle}        {Offline Handwritten Amharic Character Recognition Using Few-shot Learning}
\renewcommand{\docheader}       {M. Samuel et al.: Handwritten Amharic OCR Using Few-shot Learning} 
\renewcommand{\docauthor}       {Mesay Samuel\textsuperscript{1}, 
	Lars Schmidt-Thieme\textsuperscript{2},
	DP Sharma\textsuperscript{3},
	Abiot Sinamo\textsuperscript{4},
	Abey Bruck\textsuperscript{1}}
\renewcommand{\docaffiliation}  {\textsuperscript{1}Arba Minch University, Faculty of Computing and Software Engineering, Ethiopia,
	\email{\{mesay.samuel,abey.bruck\}@amu.edu.et},              \\
	\textsuperscript{2}Information Systems and Machine Learning Lab, 31141 Hildesheim, Germany,
	\email{schmidt-thieme@ismll.uni-hildesheim.de},      \\
	\textsuperscript{3}AMUIT MOEFDRE under UNDP,
	\email{dp.shiv08@gmail.com},
	\\
	\textsuperscript{4}Ministry of Innovation and Technology, Federal Democratic Republic of Ethiopia,
	\email{abiotsinamo35@gmail.com}}

\begin{document}
	

	
	\title{\doctitle}{\docauthor}{\docaffiliation}

	
	\setlength\columnsep{1cm}
	
	\begin{multicols}{2}
		
		{\parindent 3ex 
			\textbf{\textsf{Abstract}}---Few-shot learning is an important, but challenging problem of machine learning aimed at learning from only fewer labeled training examples. It has become an active area of research due to deep learning requiring huge amounts of labeled dataset, which is not feasible in the real world. Learning from a few examples is also an important attempt towards learning like humans. Few-shot learning has proven a very good promise in different areas of machine learning applications, particularly in image classification. As it is a recent technique, most researchers focus on understanding and solving the issues related to its concept by focusing only on common image datasets like Mini-ImageNet and Omniglot. Few-shot learning also opens an opportunity to address low resource languages like Amharic. In this study, offline handwritten Amharic character recognition using few-shot learning is addressed. Particularly, prototypical networks, the popular and simpler type of few-shot learning, is implemented as a baseline. Using the opportunities explored in the nature of Amharic alphabet having row-wise and column-wise similarities, a novel way of augmenting the training episodes is proposed. The experimental results show that the proposed method outperformed the baseline method. This study has implemented few-shot learning for Amharic characters for the first time. More importantly, the findings of the study open new ways of examining the influence of training episodes in few-shot learning, which is one of the important issues that needs exploration. The datasets used for this study are collected from native Amharic language writers using an Android App developed as a part of this study.
			
			}

		
		\section{Introduction}
		\label{sec:intro}
		Few-shot learning can be defined as a type of machine learning which aims at gaining good learning performance with few (usually 20 or less) supervised training examples \cite{wang2020generalizing}. It is an important area of machine learning which contributes to the advancement in AI in the aspect of learning humanly as humans easily learn from fewer examples. It also helps in learning rare cases which can be applied to fraud detection in electronic transactions. Few-shot learning can be applied to different application domains including computer vision, robotics, natural language processing, acoustic signal processing, drug discovery, and the like. Few-shot learning also reduces the data gathering effort and computational cost associated with big datasets which are very common issues in deep learning. 
		
		Various techniques are applied to address the problem of few-shot learning. In all the cases however, there is a way to exploit prior knowledge accumulated in the data, model or algorithm of any related machine learning task. The most common and effective way is through algorithm approach particularly meta-learning. In meta-learning an attempt is to improve performance of a new task by the meta-knowledge extracted across related tasks through a meta-learner \cite{wang2020generalizing}. Hence, the formulation of tasks plays an important role in such problems. These tasks are also known as episodes having their own training and test sets. These training and test sets are also called support and query sets respectively in few-shot learning terminologies. Each task has the same number of classes (referred as ways) for the support and query sets. However, the number of examples per class in the support set only defines the shot. Hence, a 3 way 5 shot few-shot learning problem describes a task formulation with 3 classes and 5 examples per class in the support set. 
		
		Meta-learning also known as learning to learn is any type of learning based on prior learning experience with other tasks. The similarity of the tasks also plays an important role in a way that the more similar those tasks are, the more types of meta-data one can leverage, and defining task similarity remains the key challenge. Other types of learning including multi-task learning, transfer learning and ensemble learning can also be meaningfully combined with meta-learning systems. Hence, the scientific contributions in meta-learning speed up and improve the design of machine learning pipelines and also allow us to replace hand-engineered algorithms with novel approaches learned in a data-driven way \cite{vanschoren2018meta}. Likewise this study explores and opens a wide range of perspectives into examining task similarities and their related effects on specific applications of deep learning and few-shot learning.
		
		Amharic optical character recognition in general and the handwritten character recognition in particular is not a well studied area of research. The unavailability of standard public datasets make Amharic as one of the low resource languages. Even though there are limited attempts, most of these works focus on implementation of off-the-shelf inventions which are particularly designed for Latin scripts. This trend has created two interconnected problems. The first one is  associated with the limitations to fit the real problem and proposed solution. Another problem arises from overlooking the opportunities that might have emerged with any possible innovations from the exploration of specific contexts which can then be scaled up to generalized solutions \cite{yohannes2021amharic, belay2021blended, gondere2022multi, gondere2022improving}. 
		
		In this study, offline handwritten Amharic character recognition is addressed using few-shot learning for the first time. Few-shot learning is a recent and promising area of research resolving the limitations of deep learning which requires huge amounts of labeled data. Accordingly, such techniques open a way to address low resource languages like Amharic. It is also suitable to address the issue of rarely occurring characters in real life documents. More importantly in this study, training episodes are examined from the context and nature of Amharic characters, which are the core issues in few-shot learning problems. Most of the studies in few-shot learning focus on understanding the problem itself and hence are based on common image datasets like Mini-ImageNet and Omniglot. However, in this study a more realistic application of few-shot learning is presented using prototypical networks which is a popular and simpler type of few-shot learning.
		
		The challenges facing deep learning studies when it comes to low resource languages is not only unavailability of huge standard datasets but also the difficulty of training deep learning architectures considering the fact that they typically are made of way more parameters than the dataset contains \cite{hu2021leveraging}. In regard to few-shot learning also, the datasets used to assess are not challenging and realistic as compared to the progress made in the techniques and models \cite{triantafillou2019meta}. Hence in this study, a suitable few-shot learning dataset is organized for Amharic characters using an Android App developed as a part of this study. Generally, the contributions of this paper can be summarized as follows:
		\renewcommand{\theenumi}{\roman{enumi}}
		\begin{enumerate}
			\item Organized a new few-shot learning dataset for Amharic handwritten characters with the appropriate split of train, validation, and test sets.
			\item Implemented few-shot learning for Amharic handwritten characters recognition for the first time as a benchmark. 
			\item Empirically explored how training episodes affect the performance in few-shot learning with a novel contribution in the context of Amharic handwritten characters recognition.
		\end{enumerate}

		
		\section{Related Work}
		\label{sec:related}
		Different recent papers have emerged to clarify the progress in few-shot learning \cite{triantafillou2019meta, dhillon2019baseline, chen2019closer, lake2019omniglot}. These studies mainly address the problems associated with few-shot learning datasets and performance measures. Triantafillou et al. \cite{triantafillou2019meta} proposed META-DATASET: a new benchmark for training and evaluating few-shot learning models that is large-scale, consists of diverse datasets, and presents more realistic tasks. Dhillon et al. \cite{dhillon2019baseline} performed extensive studies on benchmark datasets to propose a metric that quantifies the hardness of a few-shot episode which can be used to report the performance of few-shot algorithms in a more systematic way.
		
		Wang et al. \cite{wang2020generalizing} have made a rigorous review on few-shot learning problems to formally define and construct a good taxonomy of few-shot learning problems. Authors gave a formal definition of few-shot learning which connects to the general problem of machine learning by illustrating the unreliable empirical risk minimizer. That is the core issue in few-shot learning which arises due to fewer examples identified based on error decomposition in supervised machine learning. Therefore, few-shot learning techniques should find a way to use prior knowledge accumulated in data, model, and algorithm of other related tasks. Accordingly, Wang et al. \cite{wang2020generalizing} classified few-shot learning methods by their focus on these three constituents into data (augment training dataset using prior knowledge), model (constrain hypothesis space by prior knowledge), and algorithm (alter search strategy in hypothesis space by prior knowledge).
		
		Typical examples of algorithm few-shot learning are Model-Agnostic Meta-Learning (MAML) \cite{finn2017model} and its variants like Reptile \cite{nichol2018first}. These methods learn  parameter initialization that can be fine-tuned quickly for a new task. MAML learns initialization through effective gradient steps for a new task with a small amount of training data to produce good generalization. Reptile works this by repeatedly sampling a task, training on it, and moving the initialization towards the trained weights on that task \cite{wang2020generalizing, finn2017model, nichol2018first}. Another set of Model few-shot learning methods include siamese neural networks, matching networks, and prototypical networks which are task-invariant embedding learning models \cite{wang2020generalizing}. These methods are also known as metric learning as they learn to classify new images based on their similarity to support images unlike gradient-based meta-learning which leverages gradient descent to learn commonalities among various tasks \cite{koch2015siamese, vinyals2016matching, snell2017prototypical}. Simple Neural Attentive Learner (SNAIL) is another embedding network with interleaved temporal convolution layers and attention layers which presents an alternative paradigm where a generic architecture has the capacity to learn an algorithm that exploits domain-specific task structure \cite{wang2020generalizing, mishra2017simple}. In this paper, both model and algorithm few-shot methods are exhibited due to the implementation of prototypical networks and incorporation of auxiliary task episodic training.
		
		More related papers address few-shot learning from two main perspectives which are interrelated by nature. The first one is focusing on extraction of highly discriminative features which can easily generalize classes so that very few samples would be sufficient. Another direction is looking for different possible auxiliary or related tasks that can be used to complement the main few-shot classification task. This can be done by proposing creative classification tasks that can be trained in a multi-task learning fashion \cite{mazumder2021improving}. Mazumder et al. \cite{mazumder2021improving} proposed an approach which uses self-supervised auxiliary tasks to produce highly discriminative generic features from image datasets. The auxiliary task is a two level rotation of patches including inside the image and rotation of the whole image  and assigning one out of 16 rotation classes to the modified image. When these tasks are trained simultaneously with the main classification task, the network learns high-quality generic features that help improve the few-shot classification performance. Such methods actually utilize the concept of gradient/ optimization-based meta-learning approach of few-shot learning. Accordingly, a related work by Tripathi et al. \cite{tripathi2020few} further integrated both induction and transduction into the base learner in an optimization-based meta-learning framework. On the other hand Ravi and Larochelle \cite{ravi2016optimization}, rather than training a single model over multiple episodes, introduced an LSTM meta-learner which learns to train a custom model for each few-shot episode.
		
		A simpler and more efficient approach to few-shot learning is prototypical networks which is a metric learning. The main idea is that there exists an embedding in which points cluster around a single prototype representation for each class. Hence, the networks learn a metric space in which classification can be performed by computing distances to prototype representations of each class. Prototypical networks learn a non-linear mapping of the input into an embedding space using neural networks and take a class's prototype to be the mean of its support set in the embedding space. Classification is then performed for an embedded query point by simply finding the nearest class prototype \cite{snell2017prototypical}. Building on prototypical networks, Fort \cite{fort2017gaussian} extended to Gaussian prototypical networks incorporating a Gaussian covariance matrix, where the network constructs a direction and class dependent distance metric on the embedding space, using uncertainties of individual data points as weights.

		
		\section{Methodology}
		\label{sec:method}
		
		
		\subsection{Dataset Preparation}
		\label{ssec:dataset}
		The dataset used for this study is collected using an Android app developed as part of this study. Screenshots of the app are shown in Figure~\ref{fig1}. The link to the app with a brief notice was distributed using different platforms including email, Facebook messenger, Telegram and WhatsApp to different groups of individuals who can write Amharic. For this few-shot learning study, a total of 1,325 handwritten character images were organized. That is, five images per character for 265 Amharic characters in the Amharic alphabet. The five character images are randomly selected from those written by more than 35 individuals. The images are resized to \(32\times32\) pixels and are grouped into 120, 61, and 84 unique sets of characters for train, validation, and test splits respectively. A table showing the specific characters used for these splits is available in Appendix~\ref{Appendix_A}. The dataset of this study is publicly available\footnote{https://github.com/mesaysama/amharic-handwritten-character-dataset}
		
		\figtwocol{Screenshots of the data collection app.} 
		{fig1}   
		{0.80}         
		{Mesay_Figure_1}     
		{0}            
		
		\subsection{Prototypical Networks}
		\label{ssec:protonets}
		This study implemented Prototypical Networks as a baseline. By exploiting the opportunities from the Amharic alphabet having row-wise and column-wise similarities, meaningful alterations are made in the training episodes for the proposed methods. Prototypical Networks is one of the popular metric based few-shot learning methods which classify new classes based on their similarity to a small number of examples per class. This small number of classes is the support set which is the only information used by a few-shot classification model in order to classify query images. During training, Prototypical Networks compute a prototype for each class, which is the mean of all embeddings of support images from this class. Then, each query is simply classified as the nearest prototype in the feature space, with respect to euclidean distance. Whereas the support sets have labels both during training and testing time, the query sets have labels only during training time. Hence, for each image of the query set, the aim is to predict a label from the labels present in the support set during testing time. This study used a pretrained ResNet18 as a backbone to project both support and query images into a feature space (embedding).

		
		\section{Results and Discussion}
		\label{sec:discussion}
		All the experiments in this study are implemented using Pytorch machine learning library on the Google Colab and the computing cluster of Information Systems and Machine Learning Lab (ISMLL) from the University of Hildesheim. More importantly, the implementations are based on easy few-shot learning code by Bennequin \cite{Bennequin_easyfsl}. During this study, many few-shot learning experiments were performed to explore the possible ways one can make episodic training for the case of Amharic handwritten character recognition. For instance, one can train a model with \textit{label A} and test the model with \textit{label B} which is not the case in regular machine learning practice. \textit{Label A} could be the Amharic character row label and \textit{label B} may refer to the specific character label in the Amharic Alphabet. This is possible since a few-shot learning method allows testing a new class with its own few supervised samples. Particularly, this is due to the possibility of assigning an Amharic character to multiple labels from its tabular alphabet as shown in Appendix~\ref{Appendix_A}. 
		
		Even though there are many ways one can implement few-shot learning using episodic training, this study is limited to meaningful insights and the most common settings in the literature. Accordingly, this study implemented a few-shot learning for 1,2 and 3 shots with 5 way settings only. In all the settings, the query sets are kept to hold only two samples per class. A custom data loader is used to organize and feed the model with the appropriate combination of support and query set in each task. Training episodes are set to 20,000 tasks which are randomly generated few-shot classification tasks. The model iterates over these tasks to fit and update after each task which is called episodic training. The model predicts the labels of the query set based on the information from the support set; then it compares the predicted labels to ground truth query labels, which gives us a loss value. In this study, standard cross entropy loss and Adam optimizer are used. 
		
		For validation and testing experiments, a separate set of previously unseen classes are used with 1000 tasks each. For the case of proposed methods, half of the training episodes are replaced with row label and column label based task formulations for the proposed method 1 and 2 respectively. Sample datasets showing the task formulations for the baseline and proposed methods are presented in Figure~\ref{fig2}.
		
		\figtwocol{Sample dataset from episodes/ task formulation in the baseline and proposed methods.} 
		{fig2}   
		{0.85}         
		{Mesay_Figure_2}     
		{0}            
		
		A total of nine models are experimented to compare the results between the baseline and proposed methods for the three few-shot settings. Each experiment is run three times and the average result of the accuracy scores on a test set for 1000 character label classification task is presented in Table~\ref{tab1}. As can be seen from the results table, the baseline method scored 38.10\%, 91.10\%, and 93.70\% accuracy for 1, 2, and 3 shot settings respectively. The proposed method 1 scored 75.90\%, 90.10\%, and 88.40\% for 1, 2, and 3 shot settings respectively. Likewise, the proposed method 2 scored 38.30\%, 87.00\%, and 92.90\% for 1, 2, and 3 shot settings respectively.
		
		\tabtwocol{Accuracy scores of the competing models on a test set under different few-shot settings.} 
		{tab1}           
		{-1}                    
		{1.35}                 
		{\begin{tabular}{|p{3.5cm}|p{2.5cm}|p{2.5cm}|p{2.5cm}|}
				\hline
				\headcell{Methods} 
				& \headcell{1-shot}
				& \headcell{2-shot}
				& \headcell{3-shot} \\
				\hline
				Baseline       
				& 38.10\%
				& \textbf{91.10\%}         
				& \textbf{93.70\%}      \\
				Proposed Method 1
				& \textbf{75.90\%}
				& 90.10\%          
				& 88.40\%       \\
				Proposed Method 2
				& 38.30\%
				& 87.00\%        
				& 92.90\%       \\
				\hline
		\end{tabular}}

		An important finding of this study revealed the superior performance of the proposed method 1 which uses row based episodic training. That is, 75.90\% accuracy which outperformed other methods by a significant margin in a 1-shot setting. Even though the baseline method has a better performance for the 2 and 3-shot settings, still the proposed methods showed comparable results. That is, the results of the baseline and proposed method 1 are relatively closer to each other in a 2-shot setting. Similarly, the results of the baseline and proposed method 2 are close to each other in the 3-shot setting.
		
		Another important finding from this study is that both proposed methods have their own relevance in the different few-shot settings. That is, the proposed method 1 which uses row based episodic training helps when it is very few-shot. In contrast, proposed method 2 which uses column based episodic training appears to help when the shots are increasing. This is an interesting behavior which needs further investigation to help Amharic handwritten character recognition using few-shot learning. In line with this, both the baseline and proposed method 2 have shown an increase in performance along with the increase in shots. This is an expected empirical result in few-shot learning as studied by Triantafillou et al. \cite{triantafillou2019meta} on the effects varying the number of shots and ways in different few-shot methods including prototypical networks. Another study by Dhillon et al. \cite{dhillon2019baseline} identified that using a large number of meta-training classes results in high few-shot accuracies even for large number of few-shot classes (ways). Hence, even though the effect of varying ways is not investigated in this study, the large number of Amharic characters in the alphabet and the experiments explored in this study open an interesting area of further study.
		
		The results of the proposed methods in this study have shown how episodic task formulation affects the performance of few-shot learning. This supports the emphasis given to meta-training distribution of episodes by different studies \cite{wang2020generalizing, mazumder2021improving, tripathi2020few, ravi2016optimization, hospedales2021meta}. However, as attempted to explore in this study, the proportion and the pattern of formulating the tasks from character label, row label, and col label will remain an open area of research. This is particularly relevant during the training phase for few-shot learning since the test sets contain previously unseen classes with their own labels in the support set.
		
		Finally, this study has benefited from both types of few-shot  learning approaches including the metric learning and optimization-based learning. The use of prototypical networks helps to learn an embedding space which signifies metric learning \cite{tripathi2020few}. On the other hand, the study exploited the advantage of transfer learning and multi-task learning through episode design in meta-learning \cite{hospedales2021meta} which also improves generalization by enabling the extraction of discriminative generic features \cite{snell2017prototypical, mazumder2021improving}.  
		
		
		\section{Conclusion}
		\label{sec:conclusion}
		This study addressed offline handwritten Amharic character recognition using few-shot learning for the first time. As a baseline method, this study implemented prototypical networks which is an embedding and metric based few-shot learning method. From the opportunities of Amharic alphabet having row-wise and column-wise similarities, a novel way of augmenting the training episodes is explored as a proposed method. The results of the study revealed that the proposed method outperformed the baseline method by a significant margin in a 5-way 1-shot few-shot learning setting. This study has also proven how formulation of training episodes using related auxiliary tasks could affect the performance of few-shot learning methods. The dataset prepared by this study is another important contribution for Amharic few-shot learning research by bringing a more suitable and realistic dataset which can be used by other researchers. 
		
		Even though this study experimented with few-shot learning in different settings by varying the shots, the effects of varying ways remain unexplored. Hence, future studies can focus on extending the experiments to find the most optimum few-shot learning setting. Studying the formulation of training episodes from the combination of character, row, and column labels is also an important area of research to progress in few-shot learning for Amharic handwritten character recognition.
		
		
		\markboth{References}{References}
		\footnotesize
		\renewcommand{\refname}{References}
		\bibliographystyle{english}
		\bibliography{bibliography}
		
	\end{multicols}


\small

\begin{wrapfigure}[7]{l}[0pt]{0.6in}
	\includegraphics[height=0.9in]{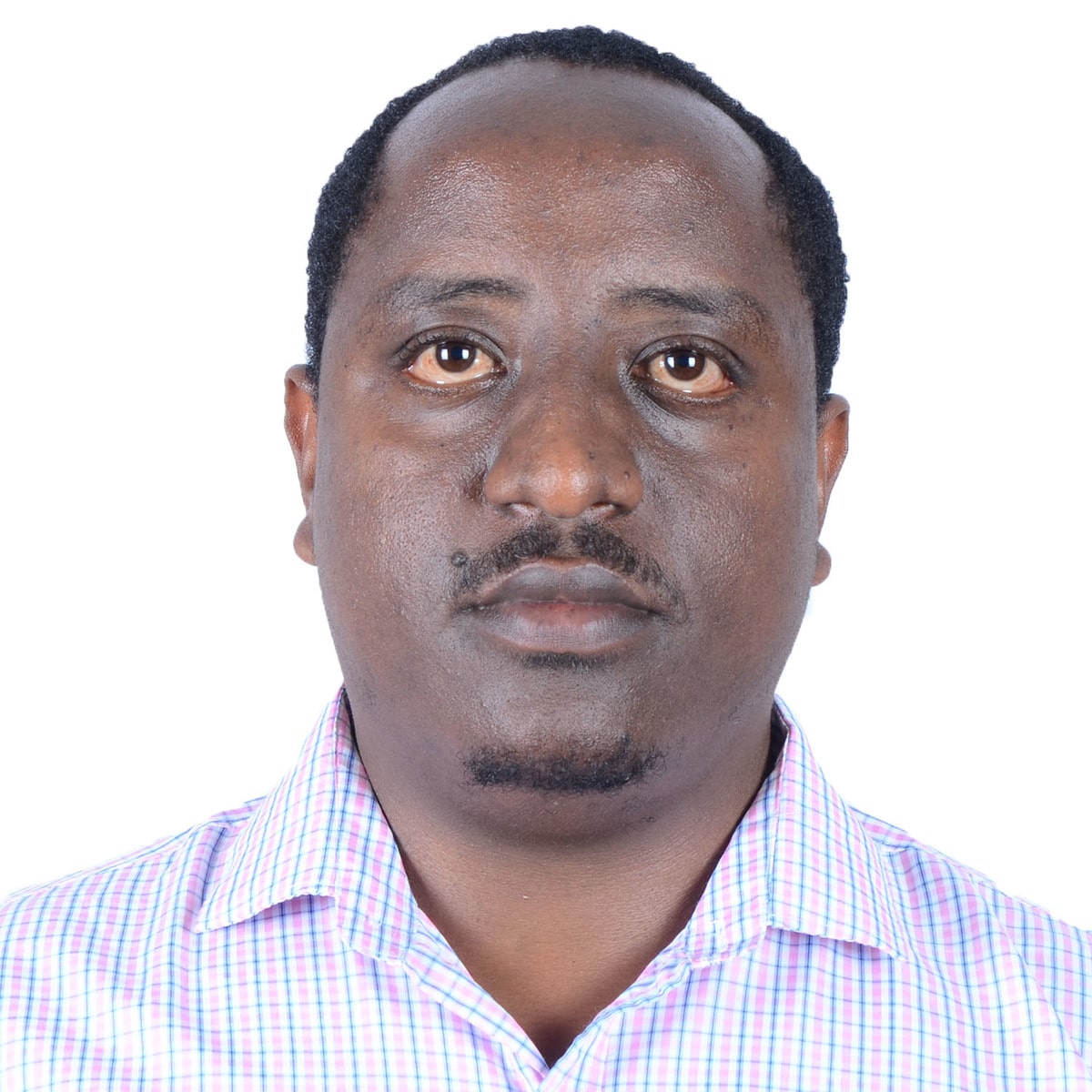}
\end{wrapfigure}
\rule{\linewidth}{0pt}
\textbf{Mesay Samuel} is a senior lecturer and PhD candidate at the faculty of Computing and Software Engineering, Arba Minch University, Ethiopia under a joint program [Arba Minch University-University of Hildesheim]. He received his MSc from Jimma University, Ethiopia in Knowledge Management (2013) and his BSc from Bahir Dar University, Ethiopia in Computer Science (2008). His current research interests include machine learning, deep learning, optical character recognition, expert systems and knowledge management.

\begin{wrapfigure}[7]{l}[0pt]{0.6in}
	\includegraphics[height=0.8in]{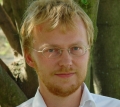}
\end{wrapfigure}
\rule{\linewidth}{0pt}
\textbf{Prof. Dr. Dr. Lars Schmidt-Thieme} is professor for Machine Learning, heading the Information Systems and Machine Learning Lab (ISMLL) at the Institute for Computer Science, University of Hildesheim since 2006. Before that, he has been assistant professor at the Institute for Computer Science at University of Freiburg from 2003 to 2006. He graduated with a PhD in Economics and Management with a thesis about frequent pattern mining in 2003 from University of Karlsruhe and with a diploma (now called master) in Mathematics in 1999 from University of Heidelberg. His research interests are supervised machine learning for complex predictors and complex decision, i.e., for all problems whose instances cannot be described naturally by a set of attributes, i.e., recommender systems, relational learning problems, time series classification etc.  

\begin{wrapfigure}[7]{l}[0pt]{0.5in}
	\includegraphics[height=0.8in]{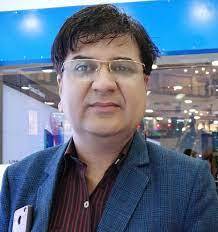}
\end{wrapfigure}
\rule{\linewidth}{0pt}
\textbf{Prof. Dr. Durga Prasad Sharma (DP Sharma)} is associated with AMUIT MOEFDRE under UNDP, MSRDCMAISM(RTU), \& Academic Ambassador, Cloud Computing (AI), IBM, USA. Prof. Sharma is a digital diplomat, computer scientist, strategic innovator, and international orator. He is the recipient of 52 National and International Awards and a wide range of appreciations including India's one of the highest civilian Awards ``Sardar Ratna Life Time Achievements International Award- 2015'' (in memory of the first Deputy Prime Minister of Independent India Sardar Vallabhbhai Patel). He has published more than a dozen books (i.e., 16 Text Books \& 6 distance education book series as writer and editor) on various themes of Computer \& IT and 131 International research papers/ articles (Print \& Digital) in refereed International Journals/conferences. He has 26 years of experience in academic, research, and professional consultancy services. He has served Mission Publiques for Internet Governance Forum under United Nations Convention. Prof Sharma has delivered 43 keynote speeches at numerous international conferences held in Canada, the USA, China, South Korea, Malaysia, India, and other countries.

\begin{wrapfigure}[7]{l}[0pt]{0.6in}
	\includegraphics[height=0.95in]{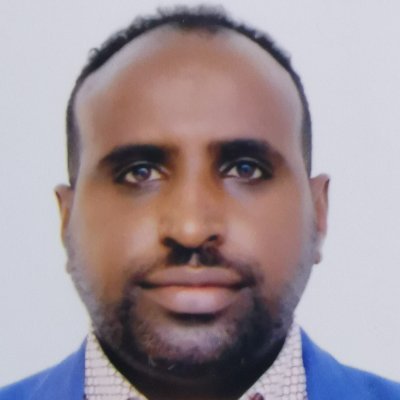}
\end{wrapfigure}
\rule{\linewidth}{0pt}
\textbf{Dr. -Ing Abiot Sinamo} is currently a Director General for ICT Sector under Ministry of Innovation and Technology of the Federal Democratic Republic of Ethiopia. He has got his PhD degree from Oldenburg University, Germany for the specializations in the areas of Intelligent Systems and ERP systems. He has assumed the post of Dean of School of Computing in Mekelle University and delivered several courses for graduate and post graduate students for a total of 17 years. He also advised number of MSc thesis works and is also co-advising two PhD works. He has published several papers in the areas of Natural Language Processing, Machine Learning, Artificial Intelligence, Knowledge Representation, Cloud Computing, Computer Vision, Software Testing, ERP Adoption, and etc in which he is interested to further his research carriers.

\begin{wrapfigure}[7]{l}[0pt]{0.6in}
	\includegraphics[height=0.9in]{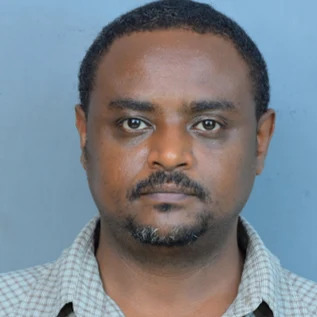}
\end{wrapfigure}
\rule{\linewidth}{0pt}
\textbf{Abey Bruck} is a lecturer at the faculty of Computing and Software Engineering, Arba Minch University, Ethiopia. He received his MSc from Addis Ababa University, Ethiopia in Information Science (2013) and his BSc from Arba Minch University, Ethiopia in Computer Science and IT (2008). His current research interests include machine learning, deep learning, and robotics.

\vfill


	\appendix
	\section{The Amharic Characters}\label{Appendix_A}
	The table below shows how the Amharic characters are selected for the train, validation, and test splits from the Amharic alphabet. This is particularly organized to be suitable for few-shot learning research. Accordingly, 120, 61, and 84 unique characters are identified for train, validation, and test sets respectively. The inclusion and exclusion of characters and their families to the specific set is done carefully so that the balance between training and evaluation challenges could not be affected.  The numbers (1-9) above the table show the column labels and the numbers (1-34) to the left show the row labels of the characters. However, the individual character label is given sequentially in the Amharic alphabet from 1 to 265. Hence, the first Amharic character for instance, has label 1 for its character, row, and column labels. Likewise, the last character has labels 265, 34, and 7 for its character, row, and column labels respectively.
	
	\begin{figure}[h]
	\includegraphics[width=8.5cm, height=24.5cm]{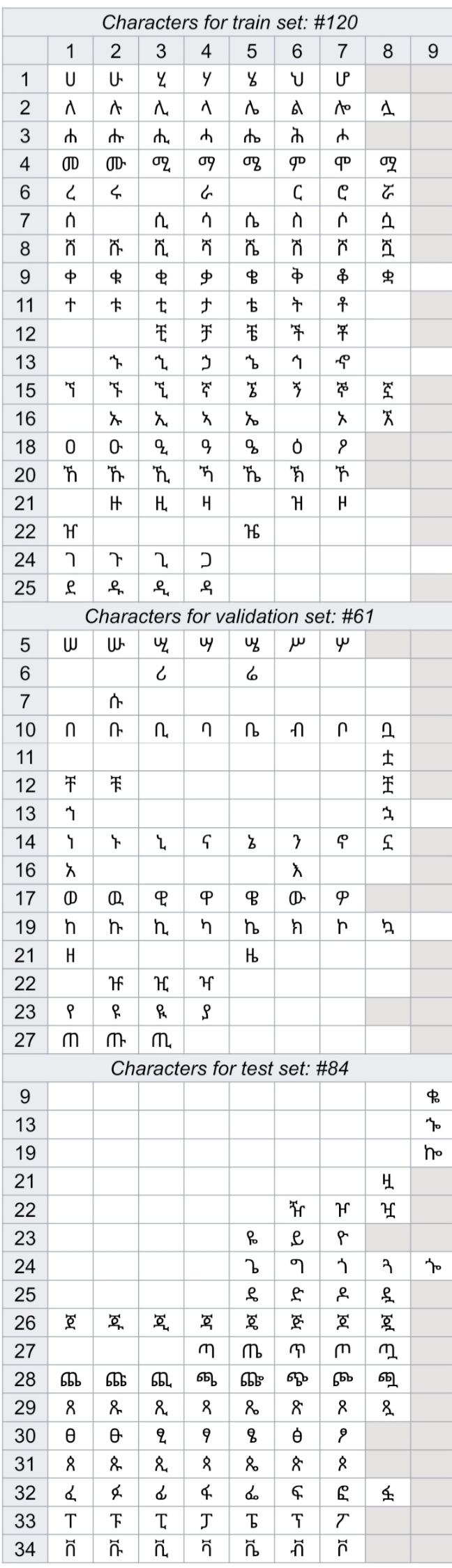}
	\end{figure}  
	

\end{document}